\documentclass{article}

\usepackage[preprint, nonatbib]{neurips_2019}

\usepackage[utf8]{inputenc} 
\usepackage[T1]{fontenc}    
\usepackage{hyperref}       
\usepackage{url}            
\usepackage{booktabs}       
\usepackage{amsfonts}       
\usepackage{amsmath}
\usepackage{nicefrac}       
\usepackage{microtype}      
\usepackage{graphicx}
\usepackage{caption}
\usepackage{subcaption}
\usepackage[ruled, vlined]{algorithm2e}  
\usepackage[sort&compress, numbers]{natbib}
\setcitestyle{square}

\newcommand{\TD}{\ensuremath{\text{TD}}}
\newcommand{\buf}{\ensuremath{\text{buf}}}

\SetStartEndCondition{ }{}{}%
\SetKwProg{Fn}{def}{\string:}{}
\SetKwFunction{Range}{range}
\SetKw{KwTo}{in}\SetKwFor{For}{for}{\string:}{}%
\SetKwIF{If}{ElseIf}{Else}{if}{:}{elif}{else:}{}%
\SetKwFor{While}{while}{:}{fintq}%

\SetKwFunction{FRecurs}{FnRecursive}%

\title{Probabilistic hypergraph grammars for efficient molecular optimization}

\author{%
  Egor Kraev \\
  Mosaic Smart Data Ltd \\
  \texttt{egor.kraev@gmail.com} \\
  \And
  Mark Harley \\
  Mosaic Smart Data Ltd \\
  \texttt{mharley.code@gmail.com}
}

\begin{document}

\maketitle

\begin{abstract}
We present an approach to make molecular optimization more efficient. We infer a hypergraph replacement grammar from the ChEMBL database, count the frequencies of particular rules being used to expand particular nonterminals in other rules, and use these as conditional priors for the policy model. Simulating random molecules from the resulting probabilistic grammar, we show that conditional priors result in a molecular distribution closer to the training set than using equal rule probabilities or unconditional priors. We then treat molecular optimization as a reinforcement learning problem, using a novel modification of the policy gradient algorithm - batch-advantage:  using individual rewards minus the batch average reward to weight the log probability loss. The reinforcement learning agent is tasked with building molecules using this grammar, with the goal of maximizing benchmark scores available from the literature. To do so, the agent has policies both to choose the next node in the graph to expand and to select the next grammar rule to apply. The policies are implemented using the Transformer architecture with the partially expanded graph as the input at each step. We show that using the empirical priors as the starting point for a policy eliminates the need for pre-training, and allows us to reach optima faster. We achieve competitive performance on common benchmarks from the literature, such as penalized logP and QED, with only hundreds of training steps on a budget GPU instance. \footnote{All code as well as the actual grammar will be made publicly available.}
\end{abstract}

\section{Introduction}\label{sec:intro}
A major problem facing the exploration of novel molecules for the purposes of drug design is the vast array of potentially useful compounds -- estimated to be in the range of $10^{24}$ and $10^{60}$ possible drug-like structures \cite{walters2019, ruddigkeit2012}. While it is of course necessary to experimentally determine the usefulness, and safety, of candidate drugs in the laboratory, de novo drug design is an approach to finding candidate molecules through either exhaustive search, or through various generative and machine learning models. This approach takes the form of an optimization procedure over given target scoring functions, giving pre-screened, promising molecules and thereby hopefully reducing drug discovery costs.

Deep learning has now been extensively investigated for encoding and generating molecular graphs \cite{duvenaud2015, kearnes2016, gilmer2017, dai2018, jin2018, simonovsky2018, kusner2017, gomezbombarelli2016, guimaraes2017, pogany2019}, and remains an area of active research. Typically, the approach taken has been to generate a linear molecular representation, such as the SMILES format \cite{weininger1988}, with an encoder-decoder network architecture similar to that used in machine translation \cite{gomezbombarelli2016}.

This route is, however, not optimal for this problem domain. Unlike written text, a molecule's structure is non-linear -- including both cycles and branches. The model is therefore forced not only to learn to optimize molecules on the given benchmark, but also to learn to generate SMILES strings corresponding to chemically valid molecules. This task is non-trivial and robs capacity of the model from the true task at hand.

A recent development which partially remedied the issue was presented by Kusner, et.~al.~ \cite{kusner2017}, in which the authors deduce a context-free grammar (CFG) for SMILES strings. This grammar guarantees that only valid SMILES strings will be produced, however not all valid SMILES strings are chemically possible molecules and so some model capacity must still be spent on learning the subset of chemically valid SMILES.

Moving away from linear representations, Kajino \cite{kajino2018} proposed the use of a grammar defined on a hypergraph representation of molecular structure. The molecular hypergraph grammar (MHG), a special case of an hyperedge replacement grammar (HRG) \cite{drewes1997}, uses rules which can be pictured as splitting the molecular graph at non-terminal bonds, and replacing them with another subgraph, thereby constructing any desired molecular structure while guaranteeing that only chemically valid molecules are produced. In particular, this approach does not have issues of invalid atom valences or loss of stereochemistry that other approaches suffer.

A different approach was taken by \cite{you2018}, who directly construct a molecular graph step by step. This has the disadvantages of a more complex action space (4 policies as compared to two in our case) as well as the fact that chemical validity must be enforced at simulation time, and the model trained to encourage valid transitions, instead of validity being guaranteed by construction as it is when using a HRG.

Brown, et.~al.~\cite{brown2019} provide an overview of the techniques used, as well as a set of benchmarks for model evaluation and the scores of several reference implementations.

We start with an approach similar to  \cite{kajino2018}, but rather than reducing all hypergraph cliques, we terminate early allowing cycles of length five and greater to remain. This permits an equally expressive grammar but allows non-trivial cycle structure to be expressed with far fewer rules, resulting in a grammar we call the hypergraph-optimized grammar (HOG). This allows the model to produce novelties without straying too far from reasonable drug-like molecules but, comes at the cost of introducing more rules. We then optimize the model's usage of the HOG by injecting conditional priors for rule selection. After parsing, we count all occurrences of rules conditional on the parent rule and the nonterminal therein that's being expanded -- counting all occurrences of $(\text{parent}, \text{location}, \text{child})$ tuples -- and use these frequencies as priors to the rule selection with the effect of grounding the molecular structure in the region of those seen in the training set, but allowing the model to explore further substructure. We infer this grammar from the ChEMBL training set of 1,273,104 SMILES strings provided by \cite{pogany2019}

The second innovation we present, first adapted to this problem in \cite{kraev2018}, is the use of the Transformer architecture \cite{vaswani2017} in place of the more typically applied RNNs \cite{yang2017, olivecrona2017} or, recently, a graph convolutional network \cite{you2018}. Unlike an RNN or CGN, the Transformer's information distance between any two inputs is always one, giving the network full information about the sequence so far when selecting the next graph node and rule to expand. This will clearly have an impact on the memory performance of the algorithm, which grows as the square of the sequence length, and so it is beneficial that the HOG optimized MHG provides a concise representation of a given molecule.

Furthermore, rather than use the encoder-decoder architecture used in some previous work \cite{dai2018, jin2018, simonovsky2018, kusner2017, gomezbombarelli2016}, we treat this problem with a reinforcement learning approach (see e.g.~\cite{popova2018} for another RL based approach) with policies selecting directly the next grammar rule and onto which hypergraph node it should be applied. This means that we do not have to learn a latent space representation of the molecules. We train these policies using a batch-advantage modification of the policy gradient algorithm.

\section{Hypergraph grammar}\label{sec:grammar}
In order to address the issue highlighted in Sec.~\ref{sec:intro} relating to SMILES string grammars, we choose to use a {\em molecular hypergraph grammar} (MHG) as derived by Kajino in \cite{kajino2018}. This works by first representing the molecular graph atoms as hyperedges and bonds as hypernodes of a molecular hypergraph. This will produce hypergraphs which are (i) 2-regular and (ii) have constrained cardinality of its hyperedges. (i) ensures that the hypergraph can be decoded back to a molecular graph and (ii) preserves the valence of each atom.

The MHG is defined over these hypergraphs as a hyperedge replacement grammar (HRG) \cite{drewes1997}, a grammar generating a hypergraph by replacement of hyperedges with other hypergraphs. This approach has a number of desirable properties, such as preserving the number of hypernodes belonging to each hyperedge, thereby preserving an atom's valence -- satisfying (i) above. Furthermore, stereochemisty can be encoded directly into the hyperedge replacement rules. MHG is thus guaranteed to produce only chemically valid molecules, allowing our model below to focus on optimizing the target benchmarks, without wasting network capacity on learning to generate valid outputs.

\subsection{Parsing algorithm}\label{sec:parsing}
\begin{figure}
    \centering
    \begin{subfigure}[b]{0.48\textwidth}
        \centering
        \includegraphics[scale=0.2]{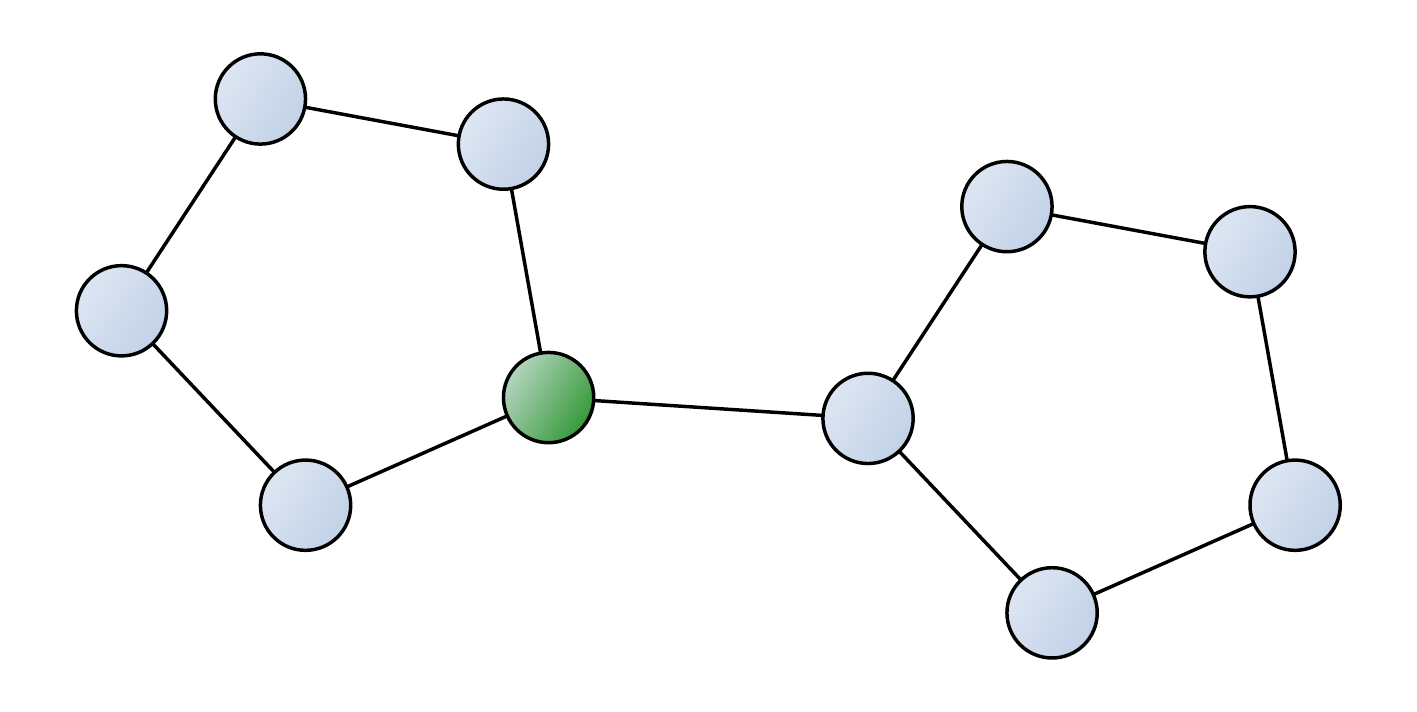}
        \caption{Molecular hypergraph before cutting.}
    \end{subfigure}
    \hfill
    \begin{subfigure}[b]{0.48\textwidth}
        \centering
        \includegraphics[scale=0.2]{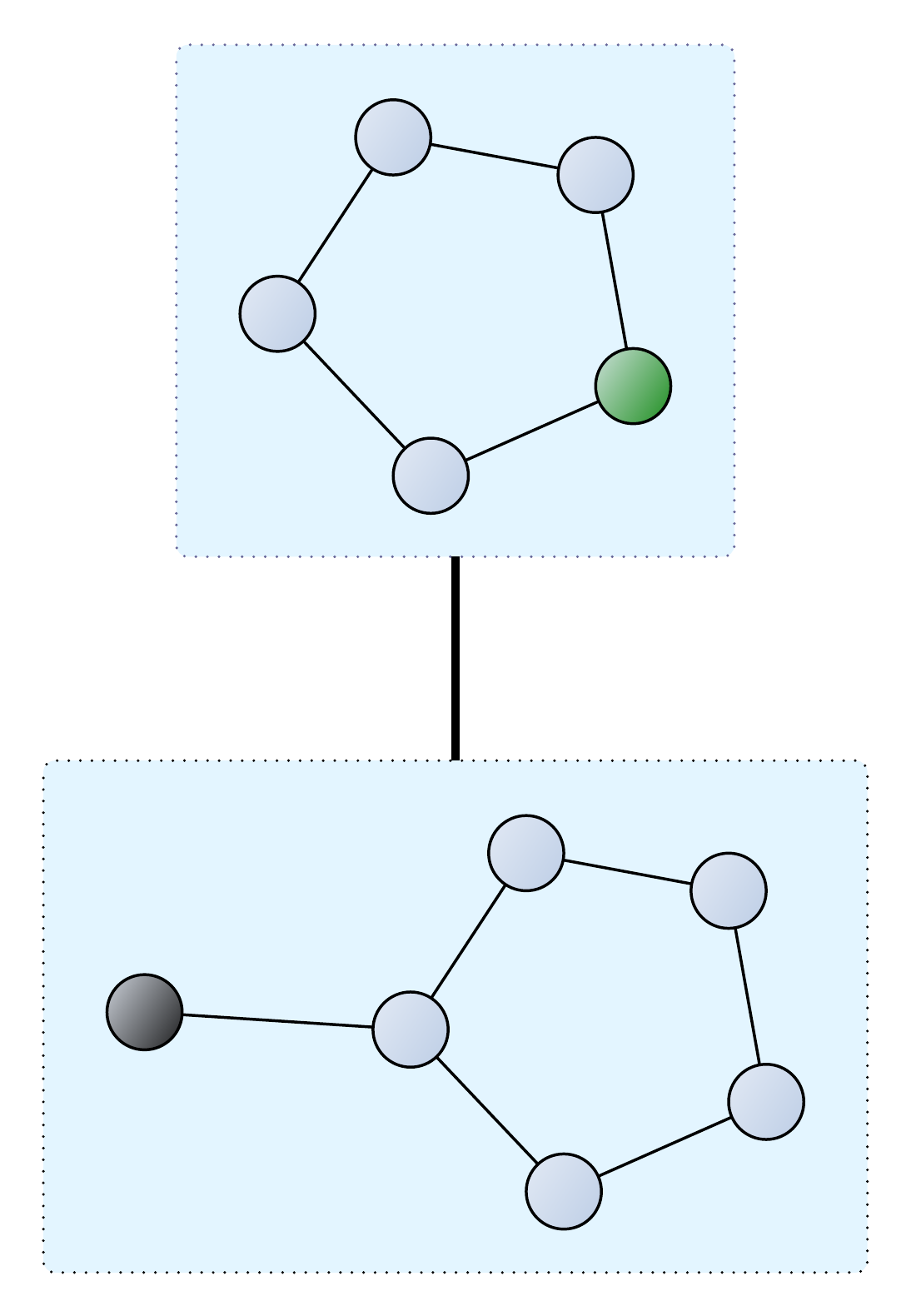}
        \caption{Parent and child sections of parse tree after cutting.}
    \end{subfigure}
    \caption{Molecular hypergraph parse tree creation step. Black node indicates the parent node that is merged into the green parent node when the child rule is applied.}
    \label{fig:parsestep}
\end{figure}

In order to build the grammar which will be used by the RL agent outlined in \ref{sec:model}, we  construct a parse tree for each molecule in a training set. By doing so for each molecule, we can identify the set of unique hyperedge replacement rules defining the grammar. Given an input molecular graph, the algorithm to deduce the MHG rules is as follows.
\begin{enumerate}
	\item Turn all atoms into child nodes containing the specific atom with a parent node containing a nonterminal with identical connectivity; this means all further graph decomposition only deals with the connectivity of the graph, without regard to the actual atoms.
	\item Find a node, $n$, at which the graph can be subdivided by cutting a single connected hyperedge
    \item Split the graph at $n$ producing a now reduced parent graph and a new child graph containing a new non-terminal node on the cut hyperedge, which we call the {\em parent node}
    \item Iterate steps above until the parent graph can no longer be subdivided in this manner
    \item Apply all of the above recursively to the new child graphs
\end{enumerate}
After this, we have a parse tree where the original graph can be reconstructed by replacing the correct node in the parent graphs with children at their corresponding parent nodes. A tree constructed in this manner will contain only graphs containing a single hyperedge or graphs composed of cycles. An example of a step in the above algorithm is illustrated in Fig.~\ref{fig:parsestep}.

\subsubsection{Extraction of Hypergraph Cliques}\label{sec:cliques}
\begin{figure}
    \centering
    \begin{subfigure}[b]{0.48\textwidth}
        \centering
        \includegraphics[scale=0.2]{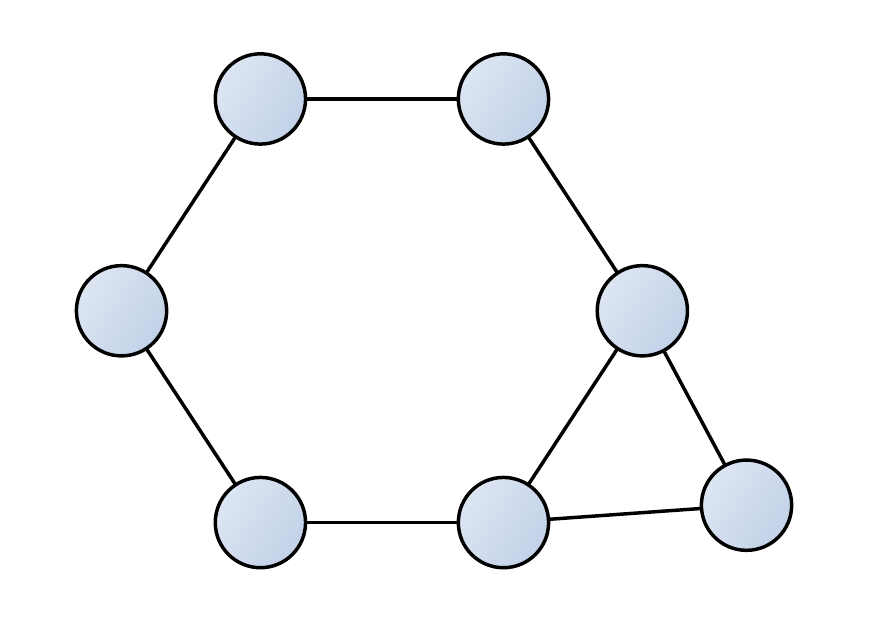}
        \caption{Molecular hypergraph before removal of the 3-clique.}
    \end{subfigure}
    \hfill
    \begin{subfigure}[b]{0.48\textwidth}
        \centering
        \includegraphics[scale=0.2]{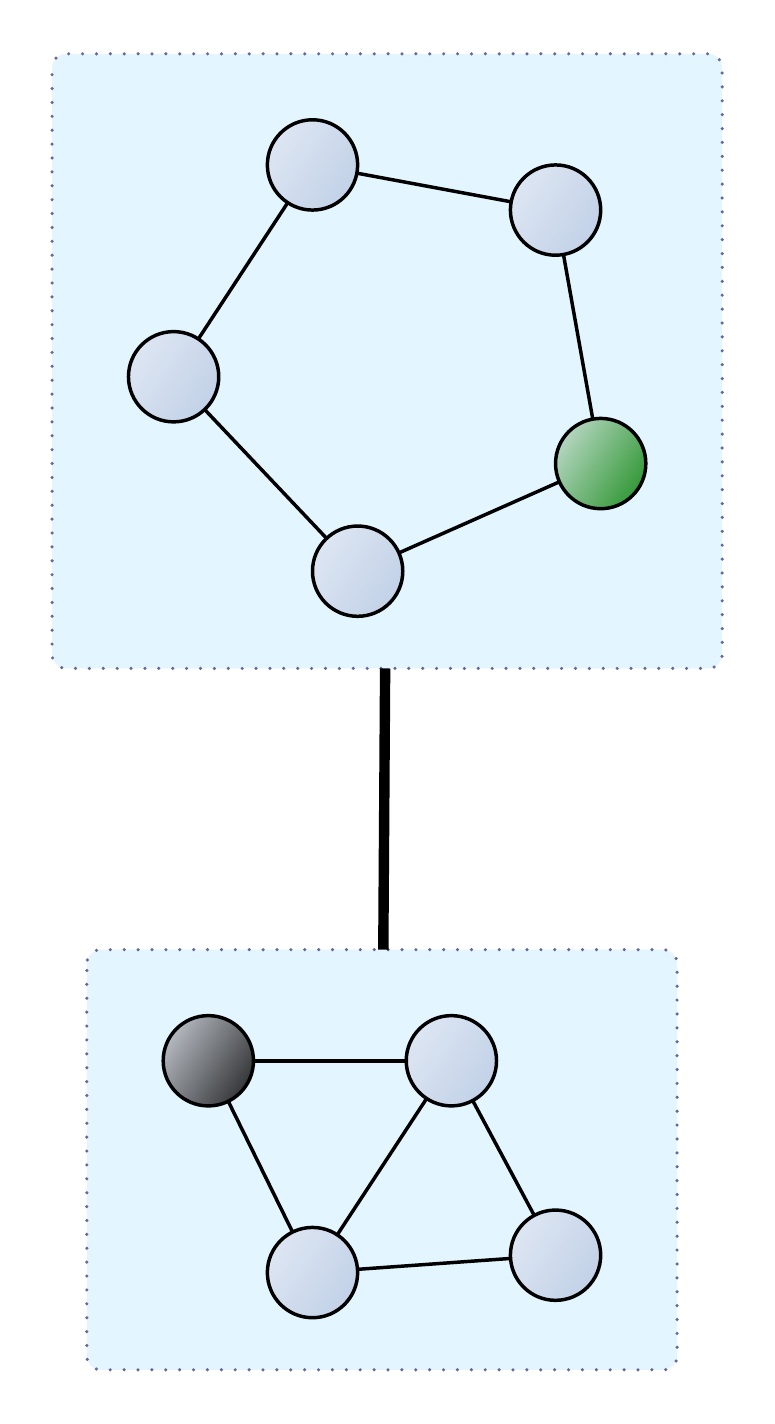}
        \caption{Parent and child sections of parse tree after removing the 3-clique.}
    \end{subfigure}
    \caption{Molecular hypergraph clique extraction step. Black node indicates the parent node that is merged into the green parent node when the child rule is applied.}
    \label{fig:cliquecollapse}
\end{figure}

Though the above procedure will produce a MHG able to represent any molecule, it is not the most efficient representation given the combinatorial complexity in the construction of molecular graph cycles leading to an unnecessarily large set of rules. We now continue to reduce the cycle containing graphs by indentifying any cliques of the remaining graph of length greater than or equal to three, but which do not contain the parent node. This is achieved by replacing the entire clique with a new non-terminal graph node in the parent, see Fig.~\ref{fig:cliquecollapse}. The child consists of the clique and all edges exiting the clique connected to the new non-terminal parent node.

After removing all such cliques, we continue to remove 2-cliques from any remaining cycles of length greater than five. This reduces the complexity of remaining cycles, and so the number of resultant rules, but discourages the production of triangles and boxes which are undesirable in output molecules from the model.

\subsection{Using the grammar to create new molecules}\label{sec:grammar_usage}
After following the procedures outlined above, we can collect all unique graphs. This set of unique graphs are the rules of the HOG. In the sense of a CFG, a rule maps a non-terminal node to a hypergraph containing further terminal and non-terminal nodes. To generate a new molecule we first select a starting rule, which by construction will contain one or more non-terminal nodes. Next, for each non-terminal node in this first graph, we select a rule whose parent node will expand the non-terminal in the parent rule. We recursively apply the same procedure now over the selected child rules until all non-terminals are replaced with terminals. This is guaranteed to produce a valid molecular hypergraph.

\subsubsection{Conditional and unconditional rule frequencies}\label{sec:freq}
To assist the model in selecting these rules, we collect the unconditional and conditional rule frequencies as they appear in the inference data set after having inferred the final set of grammar rules. Unconditional frequencies are simply the total count of appearances of each unique rule in the inference set parse trees, whereas conditional frequencies count the occurrence of each rule conditional on the given parent and the nonterminal therein, that it will expand in the parse tree. These are used as priors for rule selection, by adding their logarithm to the logits the policy model outputs, before taking the softmax to calculate rule selection probabilities.
For (parent, location, child) tuples that were never observed in the training set, we set their log prior to -3, and for those that are impossible due to nonterminal mismatch, we set their log prior to -1000; thus the priors also double as masking to ensure validity.

\subsubsection{Ensuring expansions terminate}\label{sec:termination}
The final challenge we address is making sure the rule expansion terminates before the maximum allowed number of expansion rules has been reached by the agent. To do this, we define the concept of \emph{terminal distance} of a hypergraph node, defined as the length of the shortest sequence of rules needed to expend said node into a sub-hypergraph consisting only of terminal nodes. 

We calculate this distance for each hypergraph in our set of grammar rules by means of the following dynamic programming algorithm:
\begin{enumerate}
	\item Define a set $R$ of all rule hypergraphs observed so far, and seed it with the root token \verb+molecule+. 
    \item Initialize all parent node terminal distances to $\infty$.
    \item Iterate over all parent nodes $n_p$ of rules $r$ in $R$, defining $R_{n_p}$ as the set of rules with equivalent parent node $n_p$.
	\begin{enumerate}
        \item For each $n_p$, the terminal distance is defined as one plus the minimum of terminal distances for all possible child rules of this graph. Defining C(r) as the set of child nodes of the rule $r$,
            \begin{equation}\label{eq:td}
                \TD_n(n_p) = 1 + \min_{r \in R_{n_p}}\left(\sum_{c \in C(r)} \TD_r(c) \right)\,,
            \end{equation}
            where $\TD_n, \TD_r$ are the node and rule terminal distances respectively. If the child rule, $c$, is terminal it is assigned terminal distance $\TD_r(c) = 0$.
	\end{enumerate}
\end{enumerate}
We repeat step 3.~until convergence, that is until the computed terminal distances do not change between iterations. This is most efficiently implemented as a dynamic programming problem since there are many overlapping sub-problems.

Finally, we define the terminal distance of a rule hypergraph as the sum of terminal distances of the child nodes,
\begin{equation}
    \TD_r(r) = \sum_{c \in C(r)} \TD_r(c)\,.
\end{equation}

We use the terminal distance concept to make sure the rule expansion terminates before the maximum number of steps, in the following manner: at each step, we consider the number $s$ of steps left and the terminal distance, $\text{td}$, of the sequence generated so far. 

At each step, we make sure $\text{td}\le s$, using induction. First, we choose the maximum rule sequence length to be larger than the terminal distance of the root graph. Second, at each rule selection step we consider all child rules, $r$, who can expand the next nonterminal in the graph, and only allow those where $\Delta \TD_r(r) + \text{td} \le s-1$. This must be a nonempty set because $\Delta \TD_r(r') = -1$ for at least one applicable rule $r'$. Thus, by the time we run out of steps, that is, $s=0$, we know $\text{td}=0$, that is our token sequence consists only of nonterminals. 


\section{Model choice}\label{sec:model}
\subsection{Graph embedding}
In a model that generates a discrete object via a series of decisions, we need to choose how to represent the intermediate state to feed into making the next choice. In models that produce sequences, such as SMILES strings, the natural choice is a sequence of one-hot embedded string characters. In models that construct the molecular graph directly, we have more options: we could take as our representation the sequence of the grammar rules chosen so far, like in \cite{kusner2017, kraev2018}, or a representation of the intermediate graph state, as in \cite{you2018}.

We choose the latter, via a simple encoding of three concatenated parts, with a sequence of length equaling the number of nodes in the graph. That embedding is composed of the concatenation of the connectivity matrix of the graph (with values 0, 1, 2, etc.~denoting no connection, single bond, double bond, etc.~), the identity matrix (so that the vector knows which node it refers to, as the ordering of the sequence itself is arbitrary), and a one-hot encoding of any data on each node, such as atomic number and chirality. This is then linearly transformed to the dimension of the downstream model.

\subsection{Network architecture}
We choose the Transformer architecture \cite{vaswani2017} rather than the more often used RNNs, because the nodes in the graph are not naturally ordered, and as we don't add the sinusoidal position encoding from the original Transformer model to the inputs, the outputs are invariant to the ordering of the input sequence. The Transformer was used for graph inputs by \cite{kool2018}, but never to our knowledge in the context of molecular optimization. We use model dimension 512, 6 heads, head size 64, and 5 layers.

\subsection{Policy network and discriminator}
We implement both a policy model and for some runs a discriminator\footnote{see \cite{guimaraes2017, sanchezlengeling2017, putin2018} for applications of discrimator based method, though with a sequential SMILES molecular representation} model using the above architecture. The policy model has two heads, one that supplies the logits for choosing the next nonterminal node to expand, and one that supplies the logits for the next rule to use on that node.

The discriminator network has a single head that tries to calculate the probability of whether a given molecule comes from the training dataset or has been created by the model.

\subsection{Training}\label{sec:training}
Most of the literature on molecular optimization either trains a VAE and then optimizes over the latent space, or uses a reinforcement learning approach; most of the reinforcement learning approaches don't exchange any information between the results of a simulation batch, with the exception of \cite{segler2017, neil2018} and \cite{kraev2018}, who simulate the whole model and then take the $k$ results with the best reward, and rewards the log-likelihood of those decision paths; which can be regarded as a basic kind of Monte Carlo RL. That is motivated by the fact that in contrast to many other RL use cases, we don't know the reward until the end of the episode, and so it's worth trying to reward the more successful paths as a whole after the fact.

\subsubsection{Non-originality penalty}
To avoid the optimization getting stuck in a local minimum, we penalize repeated occurrences of a molecule. Denoting the batch size used for generator training as $b$, we keep buffers of $b$, $10\cdot b$, and $100\cdot b$ last unique molecules, and decrease the rewards of new molecules if these are found in the training dataset or one of these buffers.

\subsubsection{Batch Advantage}
We extend the approach of \cite{segler2017} by introducing the concept of batch advantage, defined as the reward for a particular molecule in the batch minus the average reward for the batch. This can be regarded as a special case of the advantage concept being applied to the molecular optimization, regarding the whole sequence of decisions going into a particular molecule as one `action'. The batch reward average can then be regarded as an estimate of the state value of the initial state, and the rewards for the batch members as their exact action values. To keep the learning rate uniform, we normalize these advantage values so that the sum of their absolute values equals 1, and use them as weights for the log likelihood of the respective molecule.

Note that batch advantage is especially effective in encouraging exploration when used together with non-originality penalty. When these two are combined, if a batch returns many molecules with very similar rewards (before applying the non-originality penalty), some of them new, and some repeated, then after applying the penalty and batch advantage, the repeated molecules' log likelihood will actually have a negative weight, so we will discourage the model from reproducing these and encourage it to learn new ones.
\subsubsection{Record rewards}
As a final way to encourage exploration, we keep a buffer of the 10 best rewards observed in the current simulation. In the goal-optimization benchmarks, if a new reward is bigger than the smallest of these, we add the respective log likelihood to the objective function for that iteration, to especially encourage such molecules
\subsubsection{Training step}
A training step for the generator consists of simply computing the objective function as above and doing one step of the Adam optimizer with a learning rate of 4e-5.

For the discriminator, the key thing was to balance a quick reaction to innovations from the generator with long-term memory. To achieve that, we use the above mentioned buffers of size $b$, $10\cdot b$, and $100\cdot b$ last unique molecules, let's call them $\buf_1$, $\buf_{10}$, and $\buf_{100}$. Then each batch of $b_g$ molecules fed to the generator consists of $0.5 b_g$ molecules from the training set, $0.25 b_g$ molecules from $\buf_1$,  $0.125 b_g$ molecules from $\buf_{10}$, and  $0.125 b_g$ molecules from $\buf_{100}$. 

\section{Results}\label{sec:results}
\subsection{Comparison of distributions}
To see whether the conditional priors are a better approximation of the training set distribution, we generate sets of 10000 random molecules from our grammar, in three versions: using equal rule probabilities; using unconditional probabilities (rule frequencies); and using conditional priors as discussed in Section~\ref{sec:freq}. We further train a generator/discriminator pair, using the probability output by the discriminator as a reward for the generator, with repetition penalty and batch advantage-weighted log likelihood as generator objective (no rewards for record values in this case as we're interested in distribution as a whole, not the outliers).

The results are in Table~\ref{tab:distr}. As we see, both the KL divergence and the Frechet ChemNet distance improve as we move from random, to unconditional, to conditional priors. These are further improved by training a generator/discriminator pair, which also greatly improves uniqueness. Finally, as expected the Validity benchmark is at 1.0 throughout, as all molecules are valid by construction.

\subsection{Goal-optimization benchmarks}
We first look at the goal-optimization benchmarks called `trivial' by GuacaMol. Even though these are indeed not sufficient to measure a model's superiority for de novo drug design, they still provide an important baseline for any approach to satisfy.
Here we don't use a discriminator, do use batch advantage and repetition penalty, and also use the reward for record values as described in earlier sections.

The results are shown in Table~\ref{tab:guac-trivial}. Here, `PHG result' refers to the value computed by the GuacaMol benchmark code, which usually involves averaging over top 1, 10 and 100 molecule scores; `Top Mol Score' is the best score in the sample and `Steps to first max score' is the number of steps from a cold start needed to achieve that score. Just like the GuacaMol results, all scores are reported to 3 digits accuracy.

As Table~\ref{tab:guac-trivial} shows, we achieve competitive performance on all of these benchmarks, and notably succeed in doing so after merely hundreds of batches, each batch having size 15. In particular, we achieved the same top QED value as \cite{you2018}.

\begin{table}
    \caption{GuacaMol Distribution Benchmark Results. Compare with table 1.~from \cite{brown2019}}
    \label{tab:distr}
    \centering
    \begin{tabular}{lllll}
        \toprule
        Benchmark & No priors & Priors & Conditional Priors & With Discriminator \\
        \midrule
        Validity & 1 & 1 & 1 & 1 \\
        Uniqueness & 0.96 & 0.71 & 0.85 & 0.9987 \\
        Novelty & 0.96 & 0.7 & 0.81 & 0.9946 \\
        KL Divergence & 0.2 & 0.59 & 0.71 & 0.79 \\
        FCD & $2 \times 10^{-5}$ & 0.01 & 0.0531 & 0.118 \\
        \bottomrule
    \end{tabular}
\end{table}

\begin{table}
	\caption{GuacaMol `Trivial' Benchmark Results}
	\label{tab:guac-trivial}
	\centering
	\begin{tabular}{lllll}
		\toprule
		Benchmark & GuacaMol Best & PHG Result & Top Mol Score & \parbox{2cm}{Steps to first top score} \\
		\midrule
		$\text{logP} = -1$ & 1 & 1 & 1 & 40 \\
		$\text{logP} = 8.0$ & 1 & 1 & 1 & 130 \\
		TPSA (target 150) & 1 & 1 & 1 & 40 \\
		CNS MPO & 1 & 1 & 1 & 1 \\
		QED & 0.948 & 0.946 & 0.948 & 1200 \\
		C7H8N202 & 1 & 0.990 & 1 & 130 \\
		P. MPO & 0.998 & 0.997 & 0.999 & 780 \\
		\bottomrule
	\end{tabular}
\end{table}

For continuity with prior literature, we also ran the penalized logP optimization as optimized by \cite{kusner2017, you2018}, and many others. In this case we did use a discriminator as part of the penalty, to reward similarity to training set molecules. While our approach does not allow to directly specify maximum number of atoms, we achieved that by adding a term to the objective function penalizing number of atoms greater than a set limit. When we limit the number of atoms to 38, as in \cite{you2018}, we achieve a reward of 8.27 within 960 training steps, as always without pre-training.

Experimenting with the number of atoms limit, we found that it was the determining factor in the maximum score achieved, thus we conjecture that the increase in the penalized score that \cite{you2018} achieved (from between 5 and 6 to just under 8) was to a large extent due to a larger maximum number of atoms compared to earlier models.

\section{Discussion}
When training a model, it's best to start with the best inductive bias we can, so the model's training can focus on the metrics of interest, rather than for example chemical validity. To achieve that in the case of molecular optimization, we first constructed a hypergraph grammar for molecular optimization, by inferring it from a training dataset similarly to \cite{kajino2018}. That approach allows us to guarantee that all molecules generated by the model are chemically valid by construction, without the need for runtime checks such as done by \cite{you2018}

We then compared evenly distributed rule probability priors to those only using total rule frequency, and to conditional priors also taking into account the parent rule and the nonterminal therein that's being expanded. We showed that using these conditional priors to simulate random molecules gives us the best approximation to the training set distribution, and thus the best starting point for optimization.

Further, we introduced batch advantage, a simple and natural modification to the policy gradient approach, and showed that it encouraged exploration and sped up convergence, especially when used together with a penalty for repetition.

Having good priors makes it easier for a model to do objective optimization. That allowed us to achieve competitive results without pre-training on all the `trivial' goal-oriented GuacaMol benchmarks, in a very small number of steps, as well as on the penalized logP objective common in the literature;  our approach showed promise, but was less successful, on the more advanced GuacaMol benchmarks - we conjecture that a more sophisticated reinforcement learning algorithm, such as PPO, is needed there.

We are also, to our knowledge, the first to apply the Transformer architecture to graph embeddings for the purpose of molecular optimization. Comparison with the graph convolutional networks for that purpose will be the subject of further work.

\medskip

\small
\bibliography{nips_paper} 
\bibliographystyle{ieeetr}

\end{document}